\documentclass[runningheads]{llncs}
\usepackage{cite}
\usepackage{amsmath,amssymb,amsfonts}
\usepackage{mathtools}
\usepackage{epsfig}
\usepackage[mathscr]{euscript}
\usepackage{color}
\usepackage{booktabs} % For formal tables
\usepackage{stmaryrd} % Kruskal operator
\usepackage{bbm} % for digital payoff
\usepackage{blindtext, comment}
\usepackage{rotating} % to rotate some text
\usepackage{tikz}
\usepackage{standalone}
\usetikzlibrary{arrows, patterns}
\usetikzlibrary{decorations.pathreplacing,angles,quotes}
\usepackage{framed, standalone}
\usepackage{graphicx}
\usepackage{capt-of}
\usepackage{balance}
\usepackage{makecell} % break line in a table cell
\usepackage{afterpage}
\usepackage[linesnumbered,lined,boxed,commentsnumbered,ruled,vlined]{algorithm2e}
\usepackage{setspace}
\usepackage{textcomp}
\usepackage{xcolor}
\def\BibTeX{{\rm B\kern-.05em{\sc i\kern-.025em b}\kern-.08em
    T\kern-.1667em\lower.7ex\hbox{E}\kern-.125emX}}
\begin{document}
\title{%PHom-WAE: Persistent Homology \\ for Wasserstein Auto-Encoders \\ Applied to Credit Card Transactions
Visualization of AE's Training on Credit Card Transactions with Persistent Homology
}
\titlerunning{Persistent Homology for Wasserstein Auto-Encoders}
%
%\author{\Large \textit{Anonymous} \vspace{1.85cm}}
\author{Jeremy Charlier \inst{1,2} 
\and
Francois Petit  \inst{1}
\and
Gaston Ormazabal \inst{2}
\and
Radu State \inst{1}
\and
Jean Hilger\inst{3}}

%\authorrunning{Anonymous}
\authorrunning{J. Charlier et al.}
\institute{University of Luxembourg, L-1855 Luxembourg, Luxembourg  \\
\email{\{name.surname@\}@uni.lu}
\and
Columbia University, New York NY 10027, USA  \\
\email{\{jjc2292,gso7@\}@columbia.edu}
\and
BCEE, L-1160 Luxembourg, Luxembourg  \\
\email{j.hilger@bcee.lu}\\
}
\maketitle

\begin{abstract}
Auto-encoders are among the most popular neural network architecture for dimension reduction. 
They are composed of two parts: the encoder which maps the model distribution to a latent manifold and the decoder which maps the latent manifold to a reconstructed distribution.
However, auto-encoders are known to provoke chaotically scattered data distribution in the latent manifold resulting in an incomplete reconstructed distribution.
Current distance measures fail to detect this problem because they are not able to acknowledge the shape of the data manifolds, i.e. their topological features, and the scale at which the manifolds should be analyzed. 
We propose Persistent Homology for Wasserstein Auto-Encoders, called PHom-WAE, a new methodology to assess and measure the data distribution of a generative model.
PHom-WAE minimizes the Wasserstein distance between the true distribution and the reconstructed distribution and uses persistent homology, the study of the topological features of a space at different spatial resolutions, to compare the nature of the latent manifold and the reconstructed distribution. 
Our experiments underline the potential of persistent homology for Wasserstein Auto-Encoders in comparison to Variational Auto-Encoders, another type of generative model.
The experiments are conducted on a real-world data set particularly challenging for traditional distance measures and auto-encoders. 
PHom-WAE is the first methodology to propose a topological distance measure, the bottleneck distance, for Wasserstein Auto-Encoders used to compare decoded samples of high quality in the context of credit card transactions.

%We propose an new methodology to assess and compare the performance of auto-encoders used for dimension reduction. Relying on a topological technique called persistent homology, which captures the topological features of a space at different spatial resolutions, we are able to measure to which extent the auto-encoder is learning the topological features of the data.
%Our experiments underline the potential of our approach on a real-world data set, particularly challenging for traditional distance measures and auto-encoders.
%This is the first methodology to propose a topological distance measure, the bottleneck distance.
%We use our methodology to compare the performance of different auto-encoders on a data set of credit card transactions.
\keywords{Barcodes \and Encoding-Decoding \and Persistence Diagram.}
\end{abstract}
\section{Motivation} \label{sec::intro}
Dimension reduction techniques were initially driven by linear algebra with second order matrix decompositions such as the Singular Value Decomposition (SVD) \cite{golub1989cf} and, ultimately, with tensor decompositions \cite{carroll1970analysis,kolda2009tensor}, a higher order analogue of the matrix decompositions. Although SVD is fast enough to be applied on large data sets, it is limited to second order because it relies on matrices. While tensors are suitable for third-order experiments and above, the complexity of the decomposition and the time required for the computations become a limitation as soon as the size of the data set grows \cite{paatero1997weighted,bader2007temporal}. Hence, in recent years, several architectures for dimension reductions based on neural networks have been proposed. Dense Auto-Encoders (AE) \cite{lecun1989backpropagation,hinton1994autoencoders} are one of the most well established approach. More recently, the Variational Auto-Encoders (VAE) presented by Kingma et al. in \cite{kingma2013auto} constitute a well-known approach but they might generate poor target distribution because of the KL divergence.  \\

We recall an AE is a neural network trained to copy its input manifold to its output manifold through a hidden layer. The encoder function sends the input space to the hidden space and the decoder function brings back the hidden space to the input space. As explained in \cite{bengio2013generalized}, the points of the hidden space are chaotically scattered for most of the encoders, including the popular VAE. Even after proper training, groups of points of various sizes gather and cluster by themselves randomly in the hidden layer. Therefore, some features are missing in the reconstructed distribution $G(Z),Z\in \mathcal{Z}$. The description of the scattered points is very complex using traditional distance measures, such as the Euclidean distance, because they are not able to acknowledge the shapes of the data manifolds. However, persistent homology is specifically designed to highlight the topological features of the data \cite{chazal2017tda}. Therefore, building upon persistent homology, Wasserstein distance \cite{bousquet2017optimal} and Wasserstein Auto-Encoders (WAE) \cite{tolstikhin2017wasserstein}, our main contribution is to propose qualitative and quantitative ways to evaluate the scattered hidden space and the overall performance of AE.  \\

\begin{figure*}[b!]
	\begin{center}
	\definecolor{dtsfsf}{rgb}{0.8274509803921568,0.1843137254901961,0.1843137254901961}
\definecolor{rvwvcq}{rgb}{0.08235294117647059,0.396078431372549,0.7529411764705882}

\begin{tikzpicture}[scale=0.75, transform shape, line cap=round,line join=round,>=triangle 45,x=1cm,y=1cm]

% X rectangle
\draw [line width=0.25pt] (-7,1) rectangle (-4,3);
\fill [gray, opacity=.4] (-7,1) rectangle (-4,3);
\draw (-5.8,2.25) node[anchor=north west] {\Large{$\mathcal{X}$}};

% Z rectangle 
\draw [line width=0.25pt] (-1,1.5) rectangle (1,2.5);
\fill [gray, opacity=.4] (-1,1.5) rectangle (1,2.5);
\draw (-0.25,2.25) node[anchor=north west] {\Large{$\mathcal{Z}$}}; 

% tilde X rectangle
\draw [line width=0.25pt] (4,1) rectangle (7,3);
\fill [gray, opacity=.4] (4,1) rectangle (7,3);

% lines between X and Z rectangles
\draw [line width=0.25pt] (-4,3)-- (-1,2.5);
\draw [line width=0.25pt] (-4,1)-- (-1,1.5);

% lines between Z and tilde X rectangles
\draw [line width=0.25pt] (1,2.5)-- (4,3);
\draw [line width=0.25pt] (1,1.5)-- (4,1);

% G circle
\draw  [draw=black, line width=0.2mm, ->]  (0.5,3.025) arc (140:50:3.5);
\draw (2.8,4.75) node[anchor=north west] {$G$};

% Q circle
\draw  [draw=black, line width=0.2mm, ->]  (-5,3.5) arc (120.00:47:3.5);
\draw (-3.5,4.5) node[anchor=north west] {$Q$};

% barcode plot 
\draw [line width=0.25pt] (-1.5,-1)-- (-1.5,-4);
\draw [line width=0.25pt] (2.5,-4)-- (-1.5,-4);
\draw [line width=0.25pt] (2.5,-4)-- (2.5,-1);
\draw [line width=0.25pt] (2.5,-1)-- (-1.5,-1);
\draw [line width=0.25pt,color=dtsfsf] (-1,-1.68)-- (1.5,-1.68);
\draw [line width=0.25pt] (-1.1,-2.1)-- (-0.6,-2.1);
\draw [line width=0.25pt] (0.2,-2.3)-- (0.5,-2.3);
\draw [line width=0.25pt] (-0.5,-2.6)-- (1,-2.6);
\draw [line width=0.25pt] (-0.8,-2.8)-- (-0.3,-2.8);
\draw [line width=0.25pt] (0.5,-3)-- (1.4,-3);
\draw [line width=0.25pt] (-1.1,-3.2)-- (2.2,-3.2);

%line between the topological plots
\draw [line width=0.25pt, -triangle 60] (-2.5,-2.5)-- (-2,-2.5);
%\draw [line width=0.25pt, -triangle 60] (8.25,-2.5)-- (9.5,-2.5);
\draw (3.0,-2.25) node[anchor=north west] {\Large\textbf{$+$}};

%\draw[decoration={brace, raise=5pt, amplitude=7pt},decorate] (-7.5,-0.5) -- (2.1,-0.5);
%\draw [line width=0.25pt, -triangle 60] (2.5,-2.5)-- (3,-2.5);

% cluster plot
%\draw [line width=0.25pt] (-7.5,-1)-- (-7.5,-4);
%\draw [line width=0.25pt] (-7.5,-4)-- (-3.5,-4);
%\draw [line width=0.25pt] (-3.5,-4)-- (-3.5,-1);
%\draw [line width=0.25pt] (-3.5,-1)-- (-7.5,-1);
%
%\draw [line width=0.25pt] (-6.59,-2.06) circle (0.3cm);
%\fill [gray, opacity=.4] (-6.59,-2.06) circle (0.3cm);
%
%\draw [line width=0.25pt] (-4.55,-2.43) circle (0.3cm);
%\fill [gray, opacity=.2] (-4.55,-2.43) circle (0.3cm);
%
%\draw [line width=0.25pt] (-5.7,-3.3) circle (0.45cm);
%\fill [gray, opacity=.8] (-5.7,-3.3) circle (0.45cm);
%
%\draw [line width=0.25pt] (-6.45,-2.35)-- (-6.02,-2.96);
%\draw [line width=0.25pt] (-5.35,-3.05)-- (-4.8,-2.6);
%
%\draw [line width=0.25pt] (-5.4,-1.5) circle (0.25cm);
%\fill [gray, opacity=.2] (-5.4,-1.5) circle (0.25cm);
%
%\draw (-6.3,-1.95) -- (-5.625,-1.6);
%\draw (-5.2,-1.65) -- (-4.75,-2.2);

% circle + 3 lines to map to barcodes
\draw [line width=0.25pt, dashdotted, -triangle 60] (-5.5,0.75)-- (-5,0);
\draw [line width=0.25pt, dashdotted, -triangle 60] (3.65,0.75)-- (-4.9,0);
\draw [line width=0.25pt, dashdotted, -triangle 60] (-1,1.2)-- (-4.95,0);

% DOTTED LINES FOR TOPOLOGICAL REPRESENTATION
%\draw [line width=0.25pt, dotted, -triangle 60] (5.6,0.7)-- (5.6,0);
%\draw [line width=0.25pt, dotted, -triangle 60] (0.8,1.2)-- (3.5,0);
%\draw [line width=0.25pt, dotted, -triangle 60] (-4.0,0.7)-- (2.25,-0.5);

\draw [line width=0.25pt] (-5.65,-1.85)-- (-6.15,-1.5);
\draw [line width=0.25pt] (-6.65,-2)-- (-6.15,-3);
\draw [line width=0.25pt] (-6.15,-1.5)-- (-6.65,-2);
\draw [line width=0.25pt] (-6.15,-3)-- (-5.35,-3.4);
\draw [line width=0.25pt] (-5.35,-3.4)-- (-5.35,-3.4);
\draw [line width=0.25pt] (-5.35,-3.4)-- (-5.35,-3.4);
\draw [line width=0.25pt] (-5.35,-3.4)-- (-5.35,-3.4);
\draw [line width=0.25pt] (-5.35,-3.4)-- (-4.4,-3.3);
\draw [line width=0.25pt] (-4.4,-3.3)-- (-4.4,-3.3);

\draw [line width=0.25pt] (-3.85,-2.9)-- (-3.85,-2.9);
\draw [line width=0.25pt] (-3.85,-2.9)-- (-4.15,-2);
\draw [line width=0.25pt] (-4.15,-2)-- (-4.65,-1.5);
\draw [line width=0.25pt] (-4.65,-1.5)-- (-4.65,-1.5);
\draw [line width=0.25pt] (-4.65,-1.5)-- (-5.65,-1.85);
\draw [line width=0.25pt] (-6.65,-2)-- (-6.65,-2.5);
\draw [line width=0.25pt] (-6.15,-3)-- (-6.65,-2.5);

\draw [line width=0.25pt] (-4.4,-3.3)-- (-4.4,-3.3);
\draw [line width=0.25pt] (-4.4,-3.3)-- (-3.85,-2.9);

\draw [line width=0.25pt] (-3.61,-1.77)-- (-3.35,-2.35);
\draw [line width=0.25pt] (-3.35,-2.35)-- (-3.85,-2.9);
\draw [line width=0.25pt] (-3.35,-2.35)-- (-3.85,-2.9);
\draw [line width=0.25pt] (-3.85,-2.9)-- (-3.61,-1.77);
\draw [line width=0.25pt] (-3.61,-1.77)-- (-4.65,-1.5);
\draw [line width=0.25pt] (-4.15,-2)-- (-3.61,-1.77);
\draw [line width=0.25pt] (-4.15,-2)-- (-3.35,-2.35);
\draw [line width=0.25pt] (-5.65,-1.85)-- (-5.65,-1.5);
\draw [line width=0.25pt] (-5.65,-1.5)-- (-4.65,-1.5);
\draw [line width=0.25pt] (-5.65,-1.5)-- (-6.15,-1.5);

\draw [fill=black] (-5.65,-1.85) circle (2.5pt);
\draw [fill=black] (-6.15,-1.5) circle (2.5pt);
\draw [fill=black] (-6.65,-2) circle (2.5pt);
\draw [fill=black] (-6.15,-3) circle (2.5pt);
\draw [fill=black] (-5.35,-3.4) circle (2.5pt);
\draw [fill=black] (-5.35,-3.4) circle (2.5pt);
\draw [fill=black] (-5.35,-3.4) circle (2.5pt);
\draw [fill=black] (-5.35,-3.4) circle (2.5pt);
\draw [fill=black] (-4.4,-3.3) circle (2.5pt);
\draw [fill=black] (-4.4,-3.3) circle (2.5pt);

\draw [fill=black] (-3.85,-2.9) circle (2.5pt);
\draw [fill=black] (-3.85,-2.9) circle (2.5pt);
\draw [fill=black] (-4.15,-2) circle (2.5pt);
\draw [fill=black] (-4.65,-1.5) circle (2.5pt);
\draw [fill=black] (-4.65,-1.5) circle (2.5pt);
\draw [fill=black] (-6.65,-2.5) circle (2.5pt);

\draw [fill=black] (-4.4,-3.3) circle (2.5pt);
\draw [fill=black] (-3.61,-1.77) circle (2.5pt);
\draw [fill=black] (-3.35,-2.35) circle (2.5pt);
\draw [fill=black] (-5.65,-1.5) circle (2.5pt);

\fill[line width=0.25pt,color=gray,fill=gray,fill opacity=0.4] (-5.65,-1.85) -- (-6.15,-1.5) -- (-5.65,-1.5) -- cycle;
\fill[line width=0.25pt,color=gray,fill=gray,fill opacity=0.4] (-5.65,-1.85) -- (-4.65,-1.5) -- (-5.65,-1.5) -- cycle;
\fill[line width=0.25pt,color=gray,fill=gray,fill opacity=0.4] (-6.65,-2) -- (-6.15,-3) -- (-6.65,-2.5) -- cycle;
\fill[line width=0.25pt,color=gray,fill=gray,fill opacity=0.4] (-4.65,-1.5) -- (-3.61,-1.77)  -- (-3.35,-2.35) -- (-3.85,-2.9) -- (-3.85,-2.9) -- (-4.15,-2) -- cycle;

% frame for simplicial complex
\draw (-7,-1) node (v1) {} -- (-7,-4) -- (-3,-4) -- (-3,-1) -- cycle;

% labels
\draw (-0.6,-0.525) node[anchor=north west] {BARCODES};
\draw (-6.05,-0.2) node[anchor=north west] {FILTERED};
\draw (-7,-0.525) node[anchor=north west] {SIMPLICIAL COMPLEX};
%\draw (-8.75,-0.2) node[anchor=north west] {TOPOLOGICAL REPRESENTATION};
%\draw (-8.6,-0.55) node[anchor=north west] {OF THE SAMPLES DISTRIBUTION};
\draw (-3.65,2.25) node[anchor=north west] {ENCODING};
\draw (1.45,2.25) node[anchor=north west] {DECODING};
\draw (-7.45,3.45) node[anchor=north west] {ORIGINAL MANIFOLD};
\draw (2.85,3.45) node[anchor=north west] {RECONSTRUCTED MANIFOLD};
\draw (-1.825,3.05) node[anchor=north west] {LATENT MANIFOLD};
\draw (4.2,-2.05) node[anchor=north west] {BOTTLENECK};
\draw (4.575,-2.45) node[anchor=north west] {DISTANCE};
\draw [line width=0.25pt] (4,-2) rectangle (7,-3);

\draw (-7.05,3) node[anchor=north west] {$P_X$};
\draw (-1.075,2.525) node[anchor=north west] {$P_Z$}; 
\draw (3.95,3) node[anchor=north west] {$P_G$};
\draw (5.25,2.3) node[anchor=north west] {\Large{$\mathcal{X}$}};

\end{tikzpicture}
    \caption{In PHom-WAE, the function $W_c(P_X, P_G)$ between the true data distribution $P_X$ and the latent variable model $P_G$ of the manifold $\mathcal{X}$ is minimized. $P_G$ is specified by the prior distribution $P_Z$ of the latent manifold $\mathcal{Z}$ and the generative model $P_G(X|Z)$ with $X\in\mathcal{X}, Z\in\mathcal{Z}$. 
    Then, the samples $X, Z$ of each distribution $P_X, P_Z$ and $P_G$ are transformed independently in a metric data set to obtain filtered simplicial complex. It leads to the construction of the persistence diagram, summarized by the barcodes, to compute the bottleneck distance used to compare homologically the respective distribution $P_X, P_Z$ and $P_G$.}
    \label{fig::autoencoder}
	\end{center}
\end{figure*}
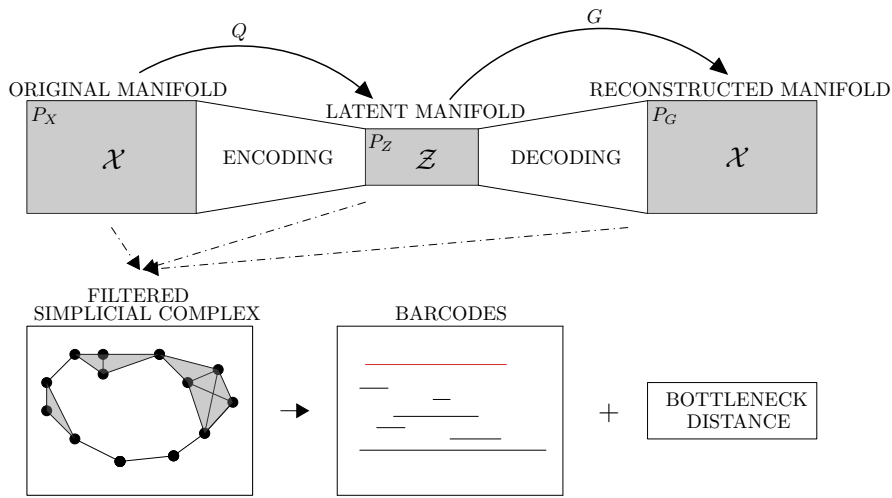

In this work we describe the persistent homology features of the encoded and decoded model distribution while minimizing the Optimal Transport (OT) function $W_c (P_X, P_G)$ for a squared cost $c(x,y)=||x-y ||_2^2$ where $P_X$ is the model distribution of the data, and $P_G$ the latent variable model specified by the prior distribution $P_Z$ of latent manifold $Z\in \mathcal{Z}$ and the generative model $P_G(X|Z)$ of the initial distribution $X\in \mathcal{X}$ given $Z$. 
The method is highlighted in figure \ref{fig::autoencoder}. 
Our contributions are summarized below:
\begin{itemize}

\item A persistent homology procedure for WAE which we call PHom-WAE to highlight the topological properties of the encoded and decoded distribution of the data for different spatial resolutions. The objective is twofold: a persistent homology description of the encoded latent space $Q_z := \mathbb{E}_{P_X}[Q(Z|X)]$, and a persistent homology description of the decoded latent space following the generative model $P_G(X|Z)$. 

\item A distance measure for persistence diagrams, the bottleneck distance, applied to WAE to compare quantitatively the true and the target distributions on any data set. We measure the shortest distance for which there exists a perfect matching between the points of the two persistence diagrams. A persistence diagram is a stable summary representation of topological features of simplicial complex, a collection of vertices, associated to the data set.

\item A persistent homology procedure to highlight the scattered latent distribution provoked by Variational Auto-Encoders (VAE) in comparison to WAE on a real-word data set. We use the concepts introduced for PHom-WAE to define PHom-VAE, Persistent Homology for VAE. We highlight the VAE's hidden layer scattered distribution using persistent homology, confirming the original statement of VAE's scattered distribution introduced in \cite{goodfellow2016deep}.

\item Finally, we propose the first application of algebraic topology and WAE on a public data set containing credit card transactions, particularly challenging for traditional distance measures and auto-encoders.
\end{itemize}

The paper is structured as follows. We discuss related works in section \ref{sec::relwork}. In section \ref{sec::propmethod}, we review the WAE formulation using OT derived by Tolstikhin et al. in \cite{tolstikhin2017wasserstein}. By using persistence homology, we are able to compare the topological properties of the distributions $P_X, P_Z$ and $P_G$, illustrated in figure \ref{fig::autoencoder}. We highlight the experimental results in section \ref{sec::exp} and we conclude in section \ref{sec::ccl} by addressing promising directions for future work.  \\

% !!! NEW SECTION !!! %
% !!! =========== !!! %
\section{Related Work} \label{sec::relwork}
\textbf{Literature on Persistence Homology and Topology}
A major trend in modern data analysis is to consider that the data has a shape, and more precisely, a topological structure. Topological Data Analysis (TDA) is a set of tools relying on computational algebraic topology allowing to obtain precise information about the data structure. Two of the most important techniques are persistence homology and the mapper algorithm \cite{chazal2017tda}.  \\

Data sets are usually represented as points cloud of Euclidean spaces. Hence, the shape of a data set depends of the scale at which it is observed. Instead of trying to find an optimal scale, persistence homology, a method of TDA, studies the changes of the topological features (number of connected components, number of holes, ...) of a data set depending of the scale. 
%The idea behind persistence homology can be traced back to the work of M. Morse in the 1940's. However, 
The foundations of persistence homology have been established in the early 2000s in \cite{frosini1999size}, \cite{Edelsbrunner2002tps}, \cite{robins2004topology} and \cite{Zomorodian2005comp}. They provide a computable algebraic invariant of filtered topological spaces (nested sequences of topological spaces which encode how the scale changes) called persistence module. This module can be decomposed into a family of intervals called persistence diagram or barcodes. This family records how the topology of the space is changing when going through the filtration \cite{ghrist2008barcodes}. The space of barcodes is measurable through the bottleneck distance. Moreover, the space of persistence module is also endowed with a metric and under a mild assumptions these two spaces are isometric \cite{Chazal2017struct}. Additionally, the \textit{Mapper} algorithm first appeared in \cite{singh2007topological}. It is a visualization algorithm which aims at produce a low dimensional representation of high-dimensional data sets in form of a graph, therefore, capturing the topological features of the points cloud.  \\

%In conjunction with these theoretical developments
Meanwhile, efficient and fast algorithms have emerged to compute persistence homology \cite{Edelsbrunner2002tps},\cite{Zomorodian2005comp} as well as to construct filtered topological spaces using, for instance, the Vietoris-Rips complex \cite{zomorodian2010fast}. Therefore, it has already found numerous successful applications. For instance, Nicolau et al. in \cite{nicolau2011topology} detected subgroups of cancers with unique mutational profile. In \cite{lum2013extracting}, it has been shown that computational topology could be used in medicine, social sciences or sport analysis. More recently, Bendich et al. improved statistical analyses of brain arteries of a population \cite{bendich2016persistent} while Xia et al. were capable of extracting molecular topological fingerprints for proteins in biological science \cite{xia2014persistent}.  \\

\textbf{Literature on Auto-Encoders and Optimal Transport} 
A large variety of AE have appeared in the last few years \cite{goodfellow2016deep}. Although promising results were achieved, most of the solutions did not address the representation of the samples of the encoded and decoded manifolds. As outlined by Bengio et al. in \cite{bengio2013generalized}, the points in the encoded manifold $\mathcal{Z}$ for the majority of the encoders are chaotically scattered. Therefore, some features are missing in the reconstructed distribution $G(Z), Z\in \mathcal{Z}$. Thus, sampling data points for the reconstruction with traditional AE is difficult. 
The added constraint of Variational Auto-Encoder (VAE) in \cite{kingma2013auto} by the mean of a KL divergence, composed of a reconstruction cost and a regularization term, provides a finer solution to generate adversarial data by reducing the impact of the chaotic scattered distribution $P_Z$ of $\mathcal{Z}$.  \\

Concurrently to the emergence of AE, Goodfellow et al. introduced the Generative Adversarial Network (GAN) model in \cite{goodfellow2014generative} in 2014. Although the GAN did not have an encoder, it consists of two parts, a generator to generate adversarial samples and a discriminator to fit the generated data points to the true data distribution. However, the GAN suffers from a mode collapse between the generator and the discriminator \cite{goodfellow2016deep}. As a solution, optimal transport \cite{villani2003topics} was applied to GAN in \cite{arjovsky2017wasserstein} with the Wasserstein distance, and therefore, introducing Wasserstein GAN (WGAN). By adding a gradient penalty to the Wasserstein distance, Gulrajani et al. in \cite{gulrajani2017improved} introduced a new training for GANs while avoiding the mode collapse. As described in \cite{chizat2015unbalanced,liero2018optimal} in the context of 
unbalanced optimal transport, Tolstikhin et al. applied these concepts to AE in \cite{tolstikhin2017wasserstein}. They proposed to add one extra divergence to the objective minimization function in the context of generative modeling leading to Wasserstein Auto-Encoders (WAE). \\

In this paper, using persistent homology and the bottleneck distance, we propose qualitative and quantitative ways to evaluate the performance of the compression of AE. We build upon the work of WAE and unbalanced OT with persistent homology. We show that that the barcodes, inherited from the persistence diagrams, are capable of representing the encoded manifold $\mathcal{Z}$ generated by the WAE. Furthermore, we show that the bottleneck distance allows to compare quantitatively the topological features between the samples $Z\in \mathcal{Z}$ of the reconstructed distribution $G(Z)$ and the samples $X \in \mathcal{X}$ of the true distribution.

% !!! NEW SECTION !!! %
% !!! =========== !!! %
\section{Proposed Method} \label{sec::propmethod}
Our method computes the persistent homology of both the latent manifold $Z\in \mathcal{Z}$ and the reconstructed manifold following the generative model $P_G(X|Z)$ based on the minimization of the optimal transport cost $W_c(P_X, P_G)$. In the resulting topological problem, the points of the manifolds are transformed to a metric space set for which a Vietoris-Rips simplicial complex filtration is applied (see definition 2). PHom-WAE achieves simultaneously two main goals: it computes the lifespan of the persistent homological features 
% birth-death of the pairing generators of the iterated inclusions to generate a topological visualization of a manifold 
while measuring the bottleneck distance between the persistence diagrams of the WAE's manifolds.

\subsection{Preliminaries and Notations}
We follow the notations used by Tolstikhin et al. in \cite{tolstikhin2017wasserstein}. Sets are denoted by calligraphic letters such as $\mathcal{X}$, random variables by capital letters $X$, and their values by lower case letters $x$. Probability distributions are denoted by capital letters $P(X)$ and their corresponding densities by lower case letters $p(x)$.

\subsection{Optimal Transport and Dual Formulation}
Following the description of the optimal transport problem \cite{villani2003topics} and relying on the Kantorovich-Rubinstein duality, the Wasserstein distance is computed as 
\begin{equation}
W_c(P_X, P_G) = 
\sup_{f \in \mathcal{F}_L} \mathbb{E}_{X\sim P_X}[f(X)]
- \mathbb{E}_{Y\sim P_G}[f(Y)]
\end{equation}
where $(\mathcal{X}, d)$ is a metric space, $\mathcal{P}(X\sim P_X, Y\sim P_G)$ is a set of all joint distributions $(X, Y)$ with marginals $P_X$ and $P_G$ respectively and $\mathcal{F}_L$ is the class of all bounded 1-Lipschitz functions on $(\mathcal{X}, d)$. \\

\subsection{Wasserstein Auto-Encoders}
As described in \cite{tolstikhin2017wasserstein}, the WAE objective function is expressed such that 
\begin{equation} \label{eq::DWAE}
  \begin{split}
    D_{\text{WAE}}(P_X, P_G) := & 
\underset{Q(Z|X)\in\mathcal{Q}}{\inf} \mathbb{E}_{P_X} \mathbb{E}_{Q(Z|X)}
[c(X, G(Z))] + \lambda \mathcal{D}_Z(Q_Z, P_Z)
  \end{split}
\end{equation}
where $c(X, G(Z)): \mathcal{X}\times \mathcal{X}\rightarrow\mathcal{R}_+$ is any measurable cost function. In our experiments, we use a square cost function $c(x,y)=||x-y||_2^2$ for data points $x,y \in \mathcal{X}$. $G(Z)$ denotes the sending of $Z$ to $X$ for a given map $G:\mathcal{Z}\rightarrow \mathcal{X}$. $Q$, and $G$, are any nonparametric set of probabilistic encoders, and decoders respectively. \\
% $Q_Z$ is the continuous mixture of the encoded data points \\
% The hyperparameter $\lambda>0$ is usually set to 10. 

We use the Maximum Mean Discrepancy (MMD) for the penalty $\mathcal{D}_Z(Q_Z, P_Z)$ for a positive-definite reproducing kernel $k:\mathcal{Z}\times\mathcal{Z}\rightarrow\mathcal{R}$
\begin{equation}
\begin{split}
  \mathcal{D}_Z(P_Z, Q_Z) := & \text{MMD}_k(P_Z, Q_Z) \\
  = & || \int_\mathcal{Z}k(z, .) dP_Z(z) - \int_\mathcal{Z}k(z, .) dQ_Z(z) ||_{\mathcal{H}_k}    
\end{split}
\end{equation}
where $\mathcal{H}_k$ is the reproducing kernel Hilbert space of real-valued functions mapping on $\mathcal{Z}$. For details on the MMD implementation, we refer to \cite{tolstikhin2017wasserstein}.  \\

\subsection{Vietoris-Rips Complex, Persistence Diagram and Barcodes}
We explain the construction of the persistence module associated to a sample of a fixed distribution on a space. First, two manifold distributions are sampled from the WAE's training. Then, we construct the persistence modules associated to each sample of the points manifolds. We refer to the first subsection of section \ref{sec::relwork} for pointed reference on persistent homology.  \\

We %first 
associate to our points manifold $\mathcal{C} \subset \mathbb{R}^n$, considered as a finite metric space, a sequence of simplicial complexes. For that aim, we use the Vietoris-Rips complex. \\

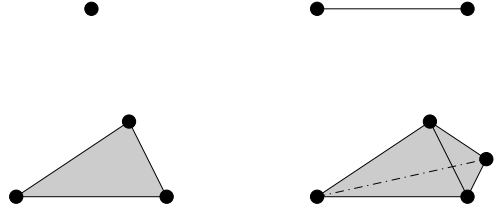
\begin{figure}[b!]
  \centering
  \definecolor{dtsfsf}{rgb}{0.8274509803921568,0.1843137254901961,0.1843137254901961}
\definecolor{rvwvcq}{rgb}{0.08235294117647059,0.396078431372549,0.7529411764705882}

\begin{tikzpicture}[scale=1, transform shape, line cap=round,line join=round,>=triangle 45,x=1cm,y=1cm]

% point
\draw [fill=black] (0,2.5) circle (2.5pt);

% line
\draw [fill=black] (3,2.5) circle (2.5pt);
\draw [fill=black] (5,2.5) circle (2.5pt);
\draw [line width=0.25pt] (3,2.5) -- (5,2.5);

% triangle
\fill [gray, opacity=.4] (-1,0) -- (1,0) -- (0.5,1) -- (-1,0);
\draw [fill=black] (-1,0) circle (2.5pt);
\draw [fill=black] (1,0) circle (2.5pt);
\draw [fill=black] (0.5,1) circle (2.5pt);
\draw [line width=0.25pt] (-1,0) -- (1,0);
\draw [line width=0.25pt] (1,0) -- (0.5,1);
\draw [line width=0.25pt] (-1,0) -- (0.5,1);

% tetrahedron
\fill [gray, opacity=.4] (3,0) -- (5,0) -- (4.5,1) -- (3,0);
\fill [gray, opacity=.4] (5,0) -- (4.5,1) -- (5.25,0.5) -- (5,0);
\draw [fill=black] (3,0) circle (2.5pt);
\draw [fill=black] (5,0) circle (2.5pt);
\draw [fill=black] (4.5,1) circle (2.5pt);
\draw [fill=black] (5.25,0.5) circle (2.5pt);
\draw [line width=0.25pt] (3,0) -- (5,0);
\draw [line width=0.25pt] (5,0) -- (4.5,1);
\draw [line width=0.25pt] (3,0) -- (4.5,1);
\draw [line width=0.25pt] (4.5,1) -- (5.25,0.5);
\draw [line width=0.25pt] (5,0) -- (5.25,0.5);
\draw [line width=0.25pt, dashdotted] (3,0) -- (5.25,0.5);

\end{tikzpicture}
  \caption{A simplical complex is a collection of numerous ``simplex" or simplices, where a 0-simplex is a point, a 1-simplex is a line segment, a 2-simplex is a triangle and a 3-simplex is a tetrahedron.}
  \label{fig::simplex}
\end{figure}

\textbf{Definition 1} Let $V=\lbrace 1, \cdots, |V|\rbrace$ be a set of vertices. A simplex $\sigma$ is a subset of vertices $\sigma \subseteq V$. A simplicial complex K on V is a collection of simplices $\lbrace \sigma \rbrace \:,\: \sigma \subseteq V$, such that $\tau \subseteq \sigma \in K \Rightarrow \tau \in K$. The dimension $n = |\sigma| - 1 $ of $\sigma$ is its number of elements minus 1. Simplicial complexes examples are represented in figure \ref{fig::simplex}.  \\

\textbf{Definition 2} Let $(X,d)$ be a metric space. The Vietoris-Rips complex at scale $\epsilon$ associated to $X$, denoted by $\text{VR}(X, \epsilon)$, is the abstract simplicial complex whose vertex set is $X$, and where $\left\lbrace x_0, x_1,...,x_k\right\rbrace$ is a $k$-simplex if and only if $d(x_i, x_j ) \leq \epsilon$ for all $0\leq i, j\leq k$. \\

We obtain an increasing sequence of Vietoris-Rips complex by considering the $\text{VR}(\mathcal{C}, \epsilon)$ for an increasing sequence $(\epsilon_i)_{1 \leq i \leq N}$ of value of the scale parameter $\epsilon$

\begin{equation}\label{diag:seqrips}
\mathcal{K}_1 \xhookrightarrow{\iota}
\mathcal{K}_2 \xhookrightarrow{\iota}
\mathcal{K}_3 \xhookrightarrow{\iota} ...
\xhookrightarrow{\iota}
\mathcal{K}_{N-1} \xhookrightarrow{\iota}
\mathcal{K}_N.
\end{equation}
Applying the \textit{k-th} singular homology functor $H_k(-,F)$ with coefficient in the field $F$ \cite{hatcher2002algebraic} to \eqref{diag:seqrips}, we obtain a sequence of $F$-vector spaces, called the \textit{k-th} persistence module of $(\mathcal{K}_i)_{1 \leq i \leq N}$

\begin{multline}
H_k(\mathcal{K}_1,F) \xrightarrow{t_1}
H_k(\mathcal{K}_2,F) \xrightarrow{t_2}
\cdots \xrightarrow{t_{N-2}} 
H_k(\mathcal{K}_{N-1},F) \xrightarrow{t_{N-1}}
H_k(\mathcal{K}_N,F).    
\end{multline}

\textbf{Definition 3} $\forall ~i<j$, the \textit{(i,j)}-persistent $k$-homology group with coefficient in $F$ of $\mathcal{K}=(\mathcal{K}_i)_{1 \leq i\leq N}$ denoted $HP_k^{i \rightarrow j}(\mathcal{K},F)$ is defined to be the image of the homomorphism $t_{j-1} \circ \ldots \circ t_i : H_k(\mathcal{K}_i,F) \rightarrow H_k(\mathcal{K}_j,F)$.\\

Using the interval decomposition theorem \cite{oudot2015Persistance}
%\cite[Theorem 1.9]{oudot2015Persistance}
, we extract a finite family of intervals of $\mathbb{R}_+$ called persistence diagram. Each interval can be considered as a point in the set $D = \left\lbrace (x, y) \in \mathbb{R}_+^2 | x \leq y \right\rbrace$. Hence, we obtain a finite subset of the set $D$. This space of finite subsets is endowed with a matching distance called the bottleneck distance and defined as follow

\begin{equation*}
d_b(A,B)=\inf_{\phi : A^\prime \to B^\prime} \sup_{ x \in A^\prime}{\lVert x-\phi(x) \rVert}
\end{equation*}
where $A^\prime= A \cup \Delta$, $B^\prime=B \cup \Delta$, $\Delta= \lbrace (x,y) \in \mathbb{R}^2_+ \vert x=y \rbrace$ and the $\inf$ is over all the bijections from $A^\prime$ to $B^\prime$.
\\

\begin{figure}[b!]
	\begin{center}
	\includegraphics[scale=0.45]{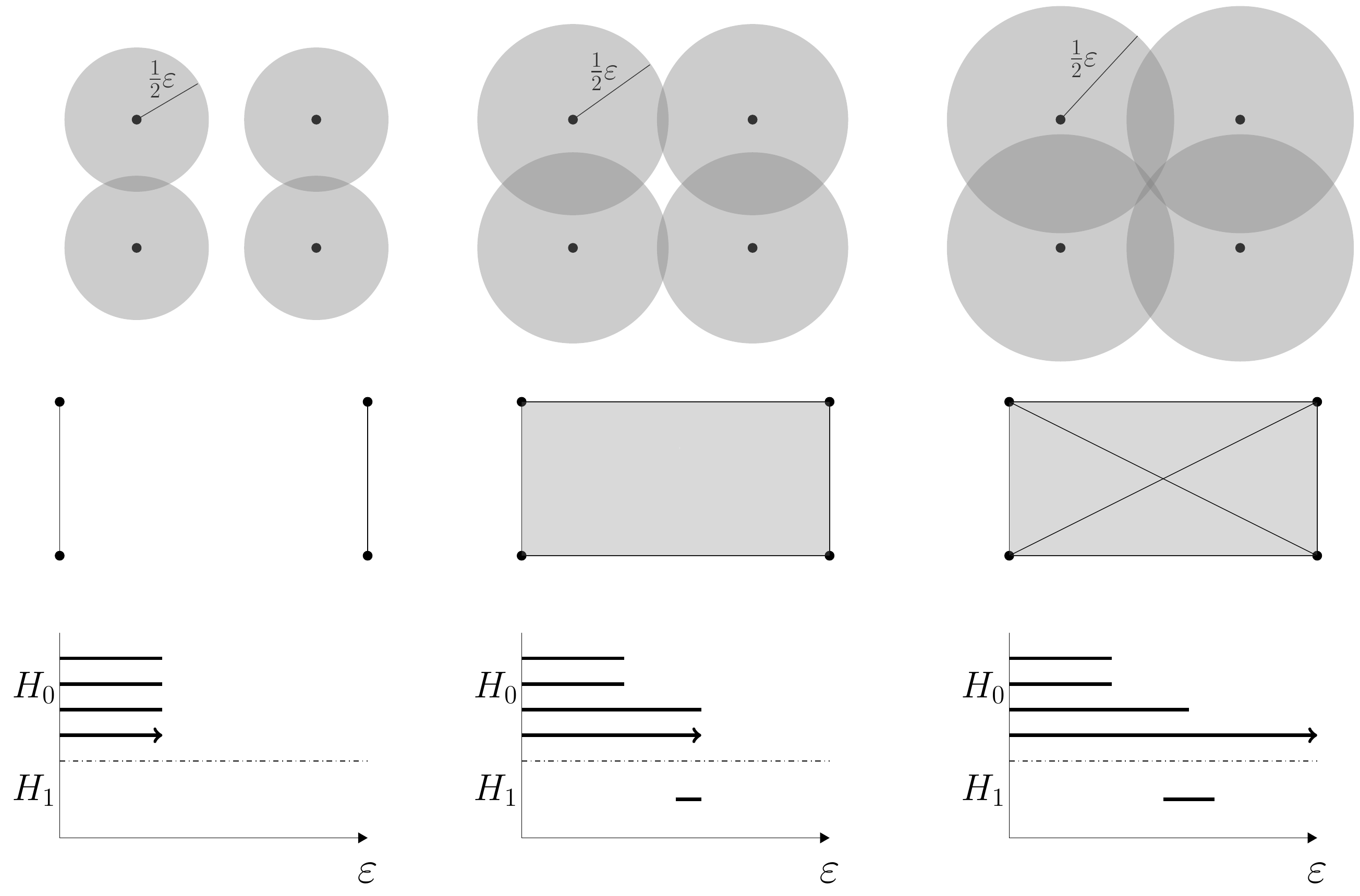}
    \caption{To describe the persistent homological features of a data set, a filtration parameter $\varepsilon$ grows around each data point, leading to geometrical disks. When two disks intersect, a line is drawn between the two original data points, creating a 1-simplex. The corresponding barcode is drawn to mark the birth-death cycle of the event according to the filtration parameter and the homological group observed. The $0$-dimensional group $H_0$ denotes the connected components and the $1$-dimensional group $H_1$ the loops.}
    \label{fig::barcode_cons}
	\end{center}
\end{figure}

\begin{figure}[t]
	\begin{center}
	\includegraphics[scale=0.6]{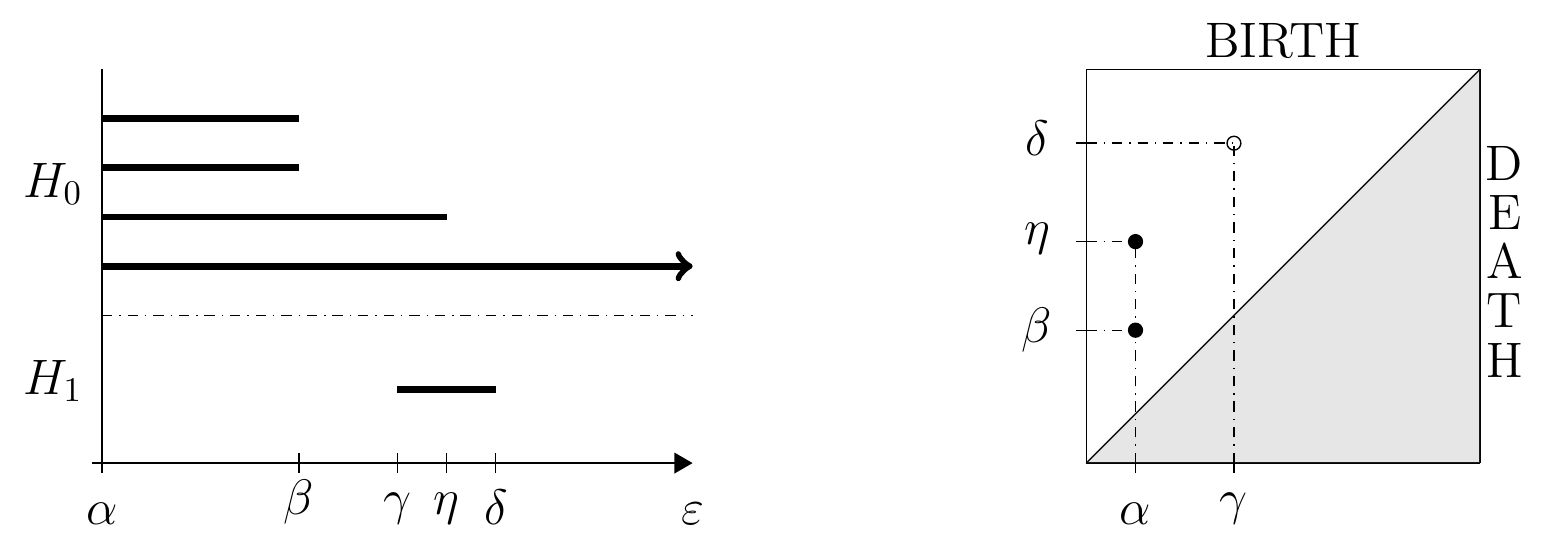}
    \caption{The filtration procedure leads to the construction of the barcodes of the homological groups $H_0$ and $H_1$, a stable summary representation of the persistence diagrams.}
    \label{fig::persdiag_cons}
	\end{center}
\end{figure}

\textbf{Application}
We illustrate the construction of the barcodes and persistence diagram with the filtration parameter $\varepsilon$ in figure \ref{fig::barcode_cons}, according to the previous definitions. For every data point, the size of the points is continuously and artificially increased using a filtration parameter $\varepsilon$. The points are, therefore, transformed to geometrical disks as the filtration parameter keeps growing. When two disks intersect, a line is drawn between the two corresponding original data points, creating a connected component defined as 1-simplex, while a barcode is drawn in a separate diagram. The barcodes highlight the birth-death cycles of each homological groups, $H_0$ for the connected components and $H_1$ for the loops. At the end of the filtration procedure, a persistence diagram is drawn to recapitulate the birth-death events observed with the barcodes, as shown in figure \ref{fig::persdiag_cons}. The persistence diagram is used to describe the topological properties of the original data points cloud using quantitative measures, such as the bottleneck distance.

\subsection{PHom-WAE, Persistent Homology for Wasserstein Auto-Encoder}
Bridging the gap between topology and neural networks, PHom-WAE uses a two-steps procedure. First, the minimization problem of WAE is solved for the encoder $Q$ and the decoder $G$. We use Adam optimizer \cite{kingma2014adam} for the optimization procedure. Then, the samples of the encoded and decoded distributions, $P_Z$ and $P_G$, are mapped to persistence homology to describe their respective manifold. \\

%After the WAE optimization, 
We highlight the topological features of the WAE's manifolds based on their respective distributions with persistent homology. First, the points contained in the manifold $\mathcal{Z}$ inherited from $P_Z$, the manifold $\mathcal{X}$ from $P_X$ and $P_G(X|Z)$ are randomly selected into respective batches. Two samples, $Y_1$ from $\mathcal{X}$ following $P_X$ and $Y_2$ from $\mathcal{X}$ following $P_G(X|Z)$, are selected to differentiate the topological features of the manifold $\mathcal{X}$ before and after the encoding-decoding. Similarly, two other samples $Y_1, Y_2$ are randomly selected from $\mathcal{Z}$ following $P_Z$ to detect the scattered distribution of the manifold $\mathcal{Z}$ after the encoding. The samples $Y_1$ and $Y_2$ are contained in the spaces $\mathcal{Y}_1$ and $\mathcal{Y}_2$, respectively. Then, the spaces $\mathcal{Y}_1$ and $\mathcal{Y}_2$ are transformed to metric space sets $\mathcal{\widehat{Y}}_1$ and $\mathcal{\widehat{Y}}_2$ for computational purposes. Then, we filter the metric space sets $\mathcal{\widehat{Y}}_1$ and $\mathcal{\widehat{Y}}_2$ using the Vietoris-Rips simplicial complex filtration. Given a line segment of length $\epsilon$, vertices between data points are created for data points respectively separated from a smaller distance than $\epsilon$. It leads to the construction of a collection of simplices resulting in Vietoris-Rips simplicial complex VR$(\mathcal{C}, \epsilon)$ filtration. In our case, we decide to use the Vietoris-Rips simplicial complex as it offers the best compromise between the filtration accuracy and the memory requirement \cite{chazal2017tda}. Subsequently, the persistence diagrams, $\text{dgm}_{Y_1}$ and $\text{dgm}_{Y_2}$, are constructed. We recall a persistence diagram is a stable summary representation of topological features of simplicial complex. The persistence diagrams allow the computation of the bottleneck distance $d_b(\text{dgm}_{Y_1}, \text{dgm}_{Y_2})$. The barcodes represent the lifespan of the homological features detected by the persistence diagrams, for instance the holes. The barcodes are a collection of the interval modules. Furthermore, the lifespan of a homological feature is defined by the boundaries of its interval module, respectively the birth time and the death time. Therefore, the barcodes illustrate in a simple way the birth-death of the pairing generators of the iterated inclusions. PHom-WAE is described in Algorithm \ref{algo::TopWAE}.

\SetAlFnt{\footnotesize}
\SetAlCapFnt{\footnotesize}
\SetAlCapNameFnt{\footnotesize}

\begin{algorithm}[b!]
\setstretch{1.25}
\DontPrintSemicolon

\KwData{training and validation sets, hyperparameter $\lambda$, encoder $Q_\phi$, decoder $G_\theta$}

\KwResult{persistent homology description of WAE's manifolds}

\Begin{

/*\textit{\small Step 1: WAE Resolution}*/

Select samples $\left\lbrace x_1, ..., x_n\right\rbrace$ from training set

Select samples $\left\lbrace z_1, ..., z_n\right\rbrace$ from validation set

Optimize until convergence $Q_\phi$ and $G_\theta$ using equation \ref{eq::DWAE} and Adam gradient descent update ($\text{lr}=0.001, \beta_1=0.9, \beta_2=0.999$)

\vspace{0.25cm}
/*\textit{\small Step 2: Persistence Diagram and Bottleneck Distance on WAE's manifolds}*/

Random selection of samples $Y_1 \in \mathcal{Y}_1, Y_2 \in \mathcal{Y}_2$ from $P_Z$ or $P_X$ and $P_G(X|Z)$

Transform $\mathcal{Y}_1$ and $\mathcal{Y}_2$ spaces into a metric space sets

Filter the metric space set with a Vietoris-Rips simplicial complex $\text{VR}(\mathcal{C}, \epsilon)$

Compute the persistence diagrams $\text{dgm}_{Y_1}$ and $\text{dgm}_{Y_2}$

Evaluate the bottleneck distance $d_b(\text{dgm}_{Y_1}, \text{dgm}_{Y_2})$

Build the barcodes with respect to $Y_1$ and $Y_2$

}
\KwRet{}

\caption{Persistent Homology for Wasserstein Auto-Encoder\label{algo::TopWAE}}
\end{algorithm}

% !!! NEW SECTION !!! %
% !!! =========== !!! %
\section{Experiments} \label{sec::exp}
In this section, we empirically evaluate the proposed methodology PHom-WAE. We assess on a %highly 
challenging data set for auto-encoders whether PHom-WAE can simultaneously achieve (i) accurate topological reconstruction of the data points and (ii) appropriate persistent homology mapping of the latent manifold.  \\

\textbf{Data Availability and Data Description}
We train PHom-WAE on one real-world open data set: the credit card transactions data set from the Kaggle database \footnote{The data set is available at https://www.kaggle.com/mlg-ulb/creditcardfraud.} containing 284\,807 transactions including 492 frauds. This data set is particularly interesting because it underlines the chaotically scattered points of the encoded manifold that are found during the AE training. Furthermore, this data set is challenging because of the strong imbalance between normal and fraudulent transactions while being of high interest for the banking industry. To preserve transactions confidentiality, each transaction is composed of 28 components obtained with PCA without any description and two additional features \textit{Time} and \textit{Amount} that remained unchanged. Each transaction is labeled as fraudulent or normal in a feature called \textit{Class} which takes a value of 1 in the case of fraud or 0 otherwise.  \\

\textbf{Experimental Setup and Code Availability}
In our experiments, we use the Euclidean latent space $\mathcal{Z} = \mathcal{R}^2$ and the square cost function $c$ previously defined as $c(x,y)=||x-y||_2^2$ for the data points $x,y \in \mathcal{X}$. The dimensions of the true data set is $\mathcal{R}^{29}$. We kept the 28 components obtained with PCA and the amount resulting in a space of dimension 29. For the error minimization process, we used Adam gradient descent \cite{kingma2014adam} with the parameters $\text{lr}=0.001, \beta_1=0.9, \beta_2=0.999$ and a batch size of 64. Different values of $\lambda$ for Wasserstein penalty have been tested, we empirically obtained the lowest error reconstruction with $\lambda=15$. The coefficients of persistence homology are evaluated within the field $\mathbb{Z}/2 \mathbb{Z}$. We only consider homology groups $H_0$ and $H_1$ who represent the connected components and the loops, respectively. Higher dimensional homology groups did not noticeably improve the results quality while leading to longer computational time. The simulations were performed on a computer with 16GB of RAM, Intel i7 CPU and a Tesla K80 GPU accelerator. To ensure the reproducibility of the experiments, the code is available at the following address\footnote{The code is available at https://github.com/dagrate/PHom-WAE.}.  \\

\textbf{Results and Discussions on PHom-WAE against PHom-VAE}
We test PHom-WAE against PHom-VAE, Persistent Homology for Variational Auto-Encoder. We recall a Variational Auto-Encoder \cite{kingma2013auto} uses a KL divergence, denoted by $D_{KL}(P_X, P_G)$, composed of a reconstruction cost and a regularization term instead of an OT cost function. We use the same concepts introduced for PHom-WAE to define PHom-VAE. We compare the performance of PHom-WAE and PHom-VAE on two specificities: the latent manifold $\mathcal{Z}$ and the reconstructed data distribution $G(Z)$ following the generative model distribution $P_G$ for the samples $Z\in\mathcal{Z}$.  \\

\begin{figure*}[t]
\centering
  \begin{turn}{90} 
   \hspace{1.6cm} Original Sample
  \end{turn}
  \frame{\includegraphics[scale=0.375]{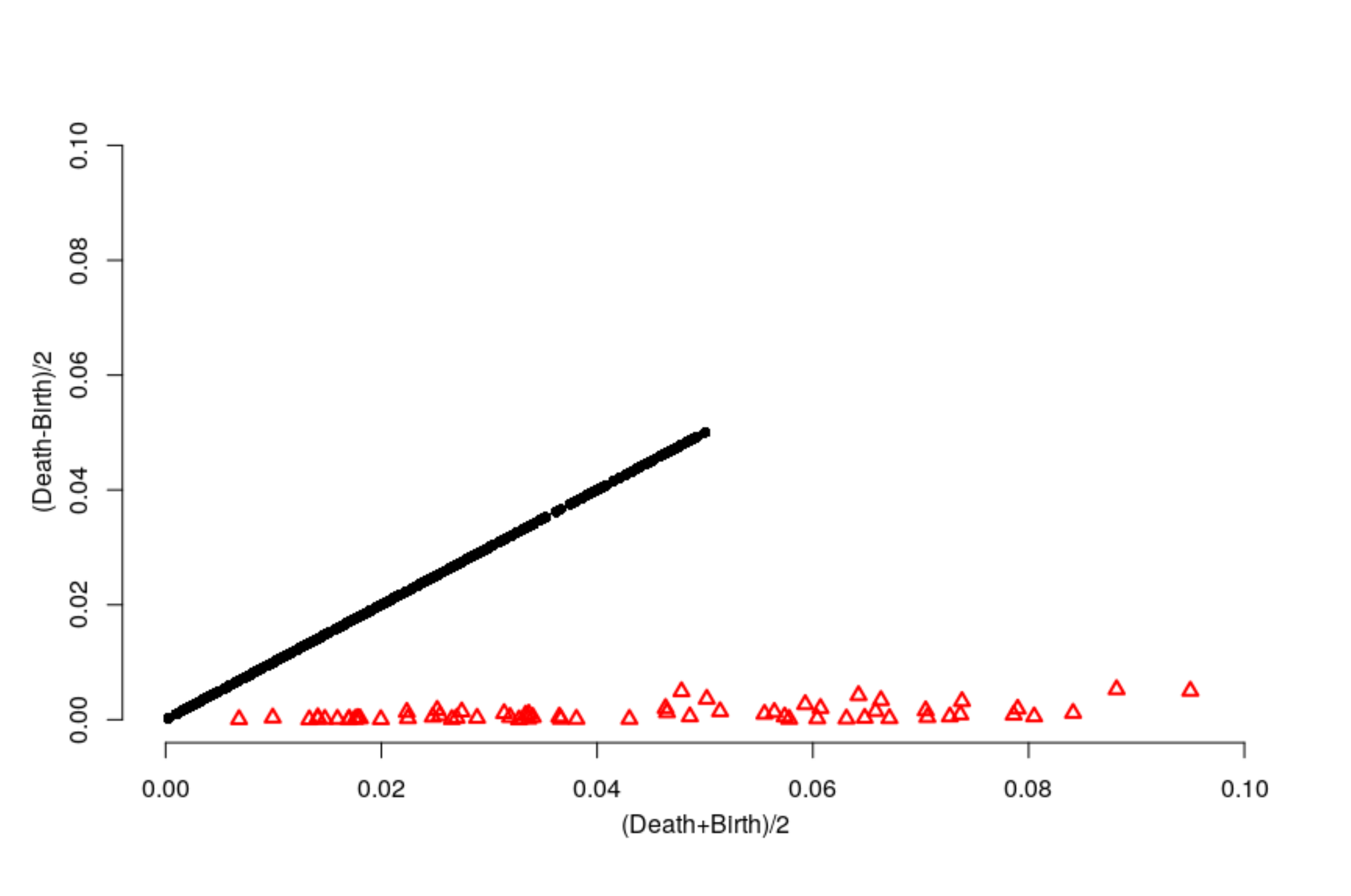}}

  \vspace{0.2cm}
  \begin{turn}{90} 
   \hspace{0.6cm} PHom-WAE
  \end{turn}
  \frame{\includegraphics[scale=0.235]{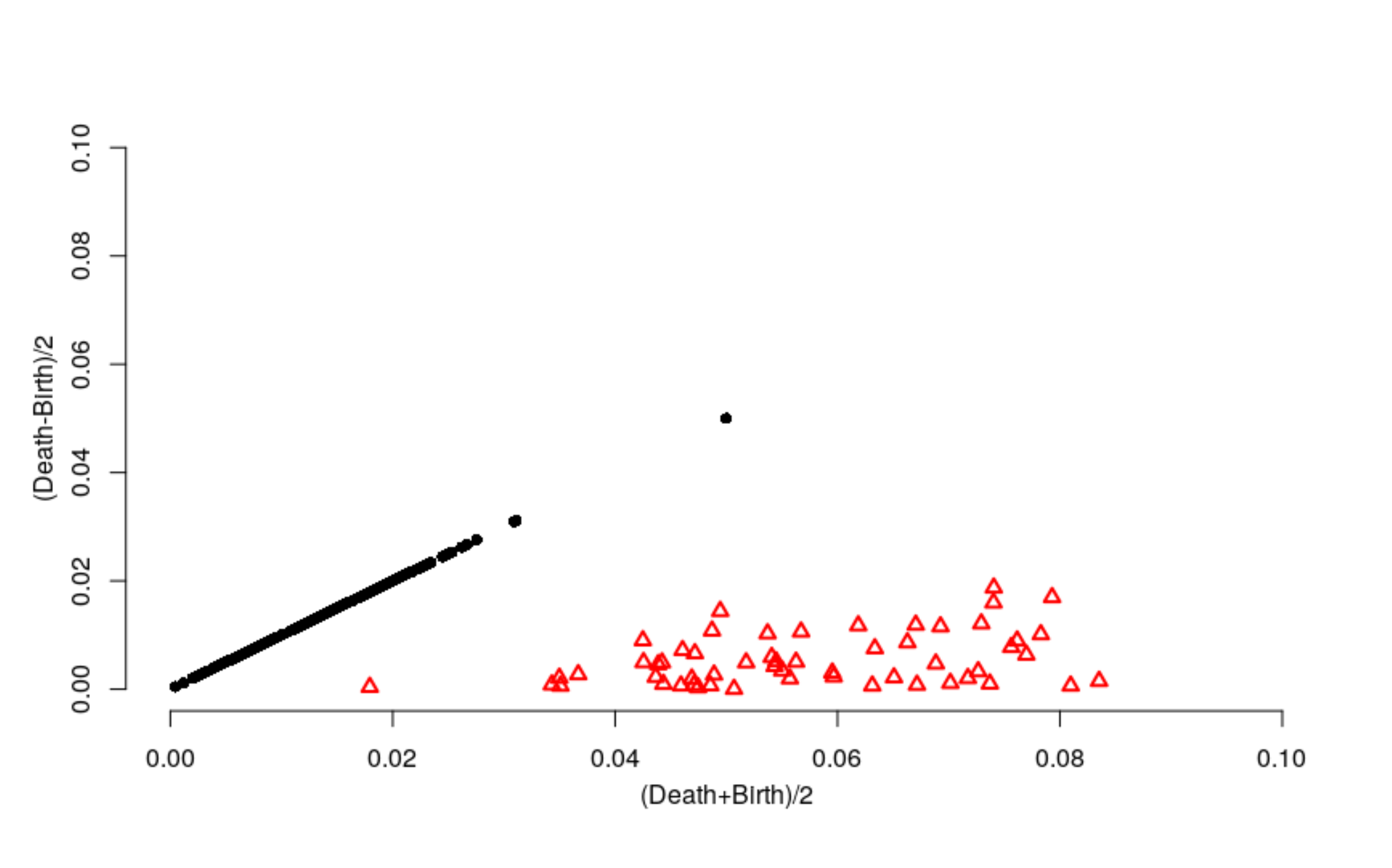}} \hspace{.3cm}
  \begin{turn}{90} 
   \hspace{0.6cm} PHom-VAE
  \end{turn}
  \frame{\includegraphics[scale=0.235]{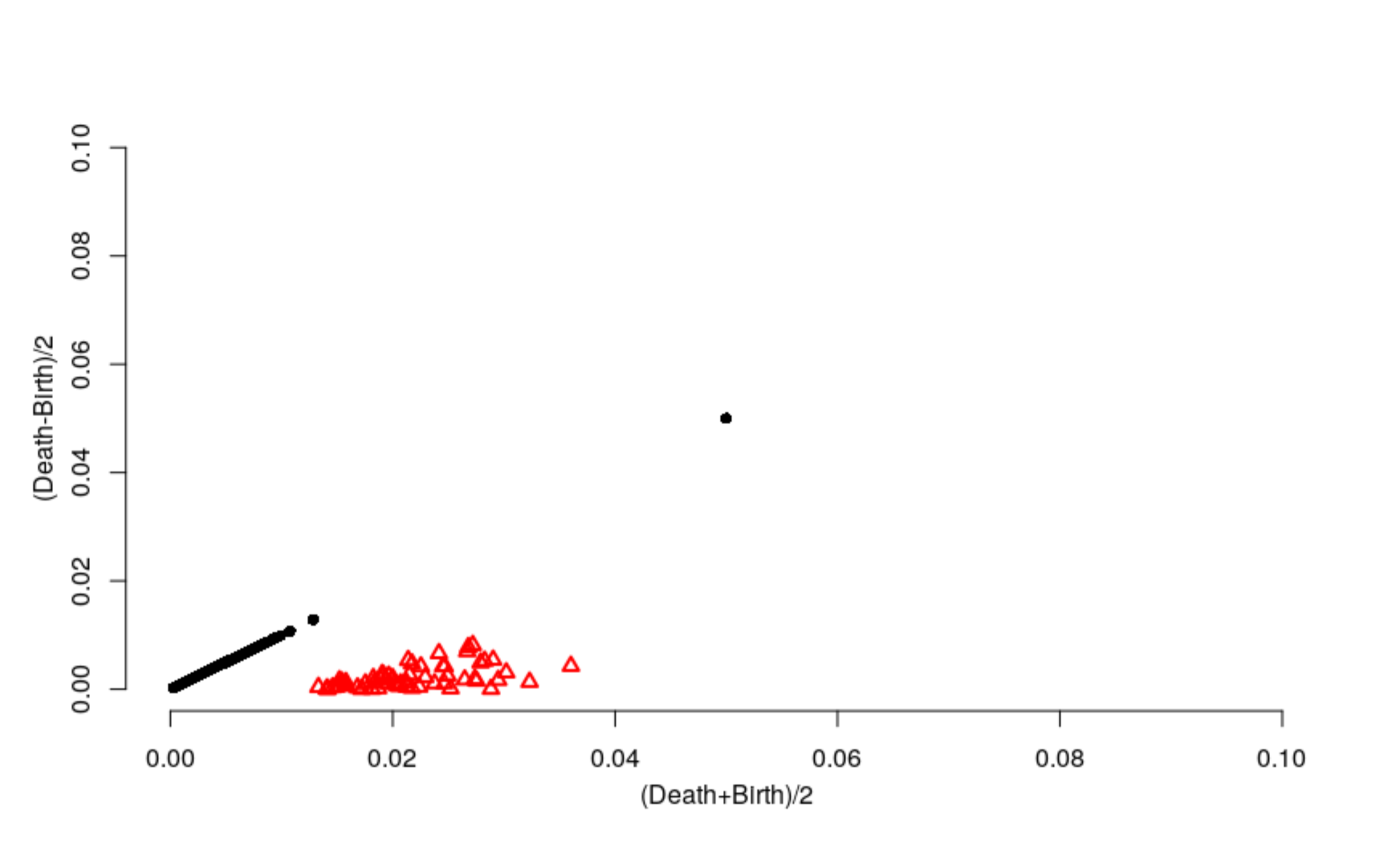}}

\caption{PHom-WAE and PHom-VAE's rotated persistence diagrams in comparison to the persistence diagram of the original sample show the birth-death of the pairing generators of the iterated inclusions. Black points represent the 0-dimensional homology groups $H_0$ that is the connected components of the complex. Red triangles represent the 1-dimensional homology group $H_1$ that is the 1-dimensional features, the cycles.} 
\label{fig::persistence_plot}
\end{figure*}

\begin{figure*}[t]
\centering
%  \begin{minipage}{1.0\textwidth}
  \centering
  \begin{turn}{90} 
   \hspace{1.75cm} Original Sample
  \end{turn}
  \frame{\includegraphics[scale=0.28]{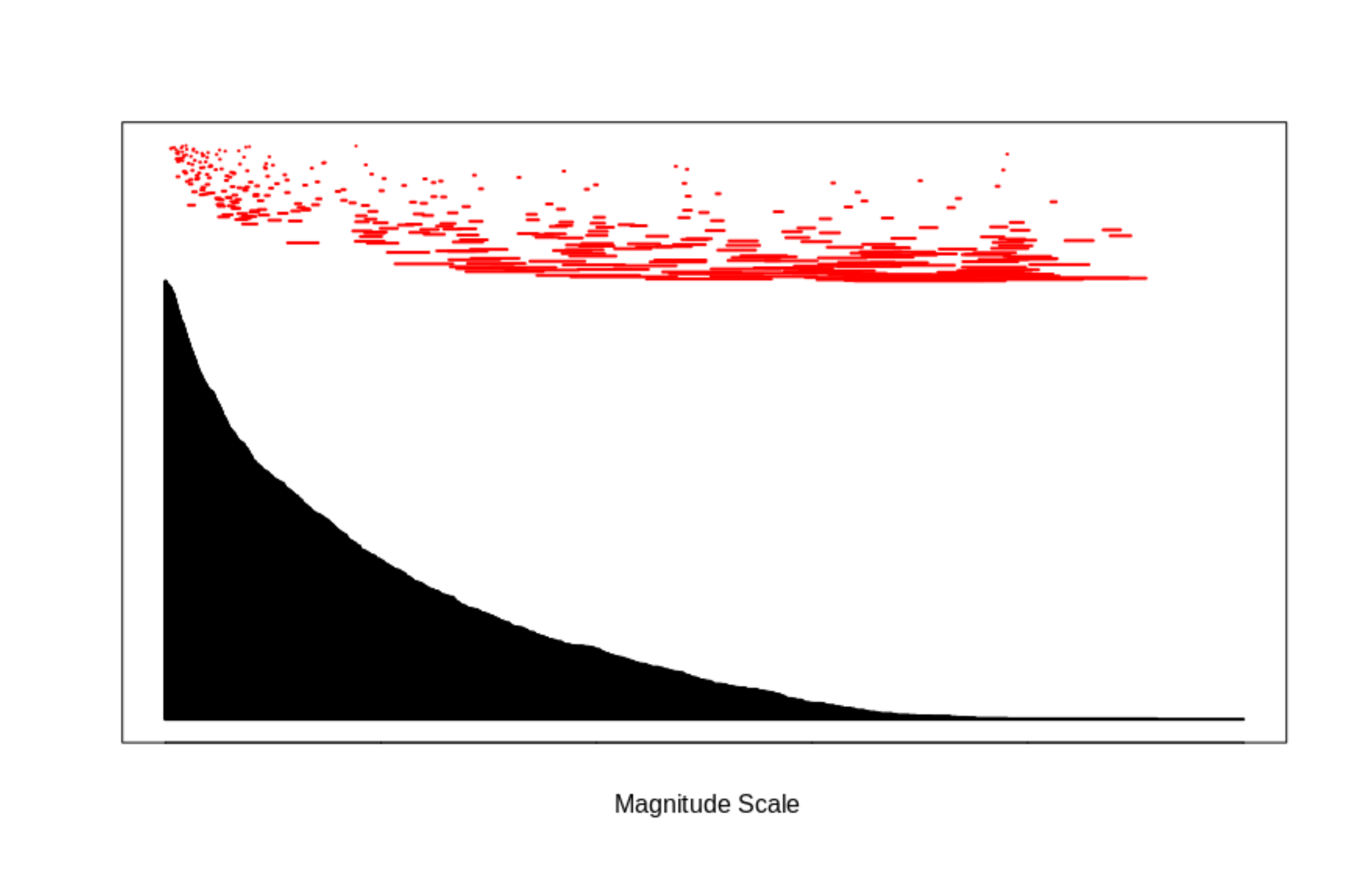}}
%  \end{minipage}\hfill
 
%  \begin{minipage}{1.0\textwidth}
%   \centering
  \vspace{0.2cm}
  \begin{turn}{90} 
   \hspace{.75cm} PHom-WAE
  \end{turn}
  \frame{\includegraphics[scale=0.18]{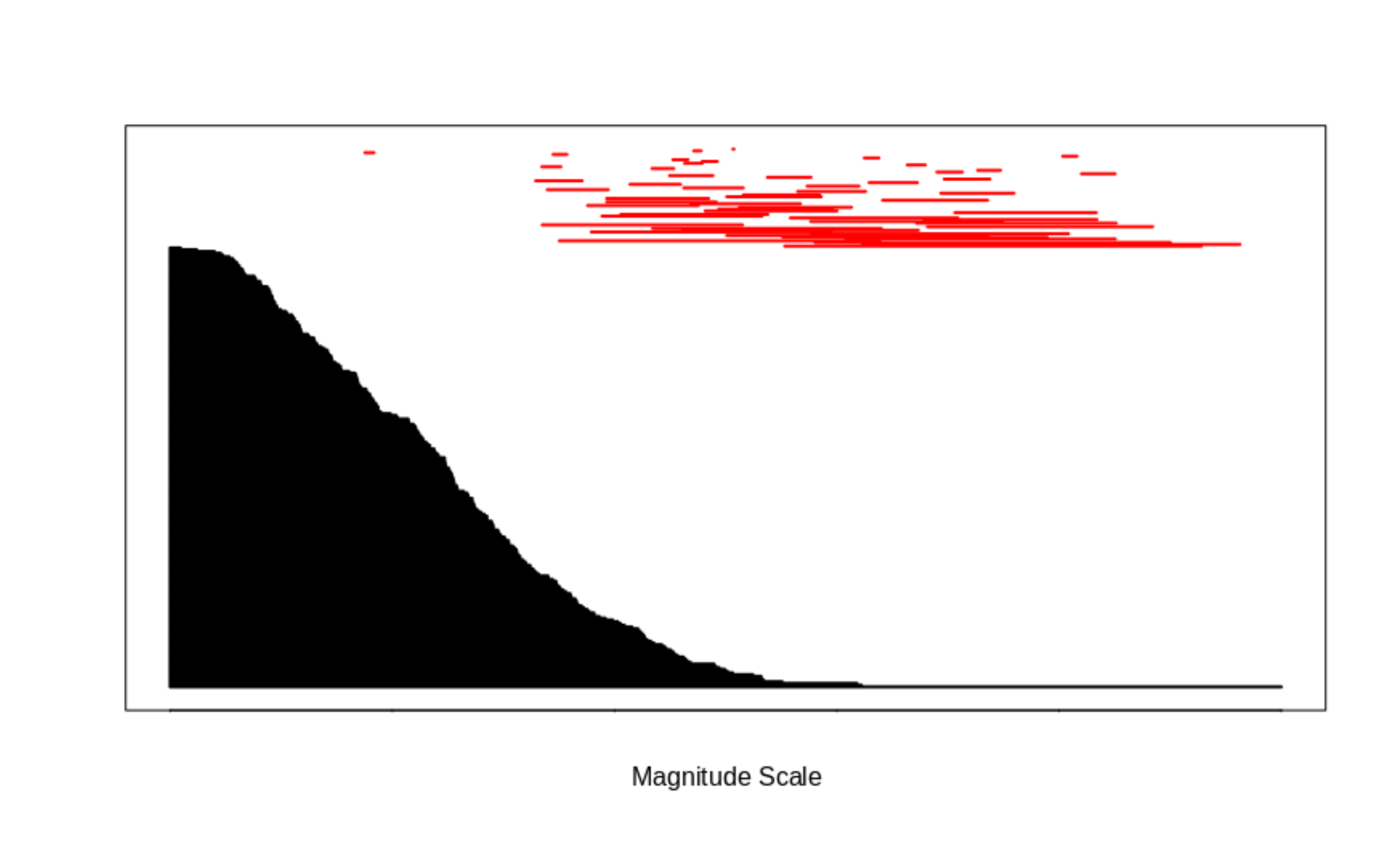}}
  \hspace{.25cm}
  \begin{turn}{90} 
  \hspace{.75cm} PHom-VAE
  \end{turn}
  \frame{\includegraphics[scale=0.18]{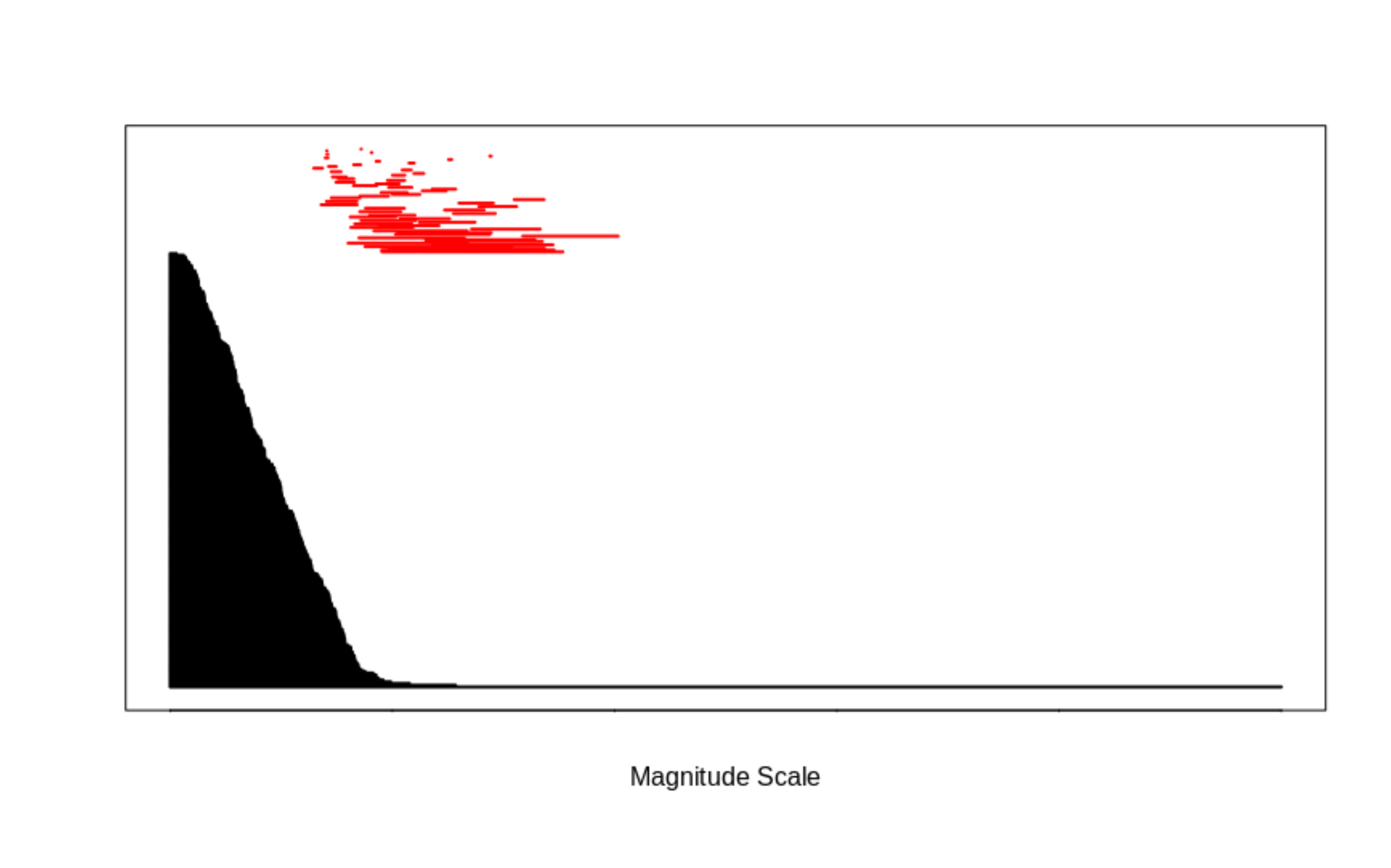}}

%  \end{minipage}\hfill
 
%  \begin{minipage}{1.0\textwidth}
%   \centering
%   \vspace{0.2cm}
%   \begin{turn}{90} 
%   \hspace{1.15cm} PHom-VAE
%   \end{turn}
%   \frame{\includegraphics[scale=0.28]{figures/vae_barcode_cs2_updated.pdf}}
%  \end{minipage}\hfill
 
\caption{PHom-WAE and PHom-VAE's barcode diagrams in comparison to the barcode diagram of the original sample based on the persistence diagrams of figure \ref{fig::persistence_plot}. The barcodes diagrams are a simple way to represent the persistence diagrams. We refer to \cite{chazal2017tda} for further details on their generation.} 
\label{fig::latentplot}
\end{figure*}

\begin{figure*}[t!]
\centering
 \begin{minipage}{.49\textwidth}
  \centering
  Original Samples \\ \vspace{0.2cm}
  \begin{turn}{90} 
   \hspace{0.5cm} PHom-WAE
  \end{turn}
  \frame{\includegraphics[scale=0.235]{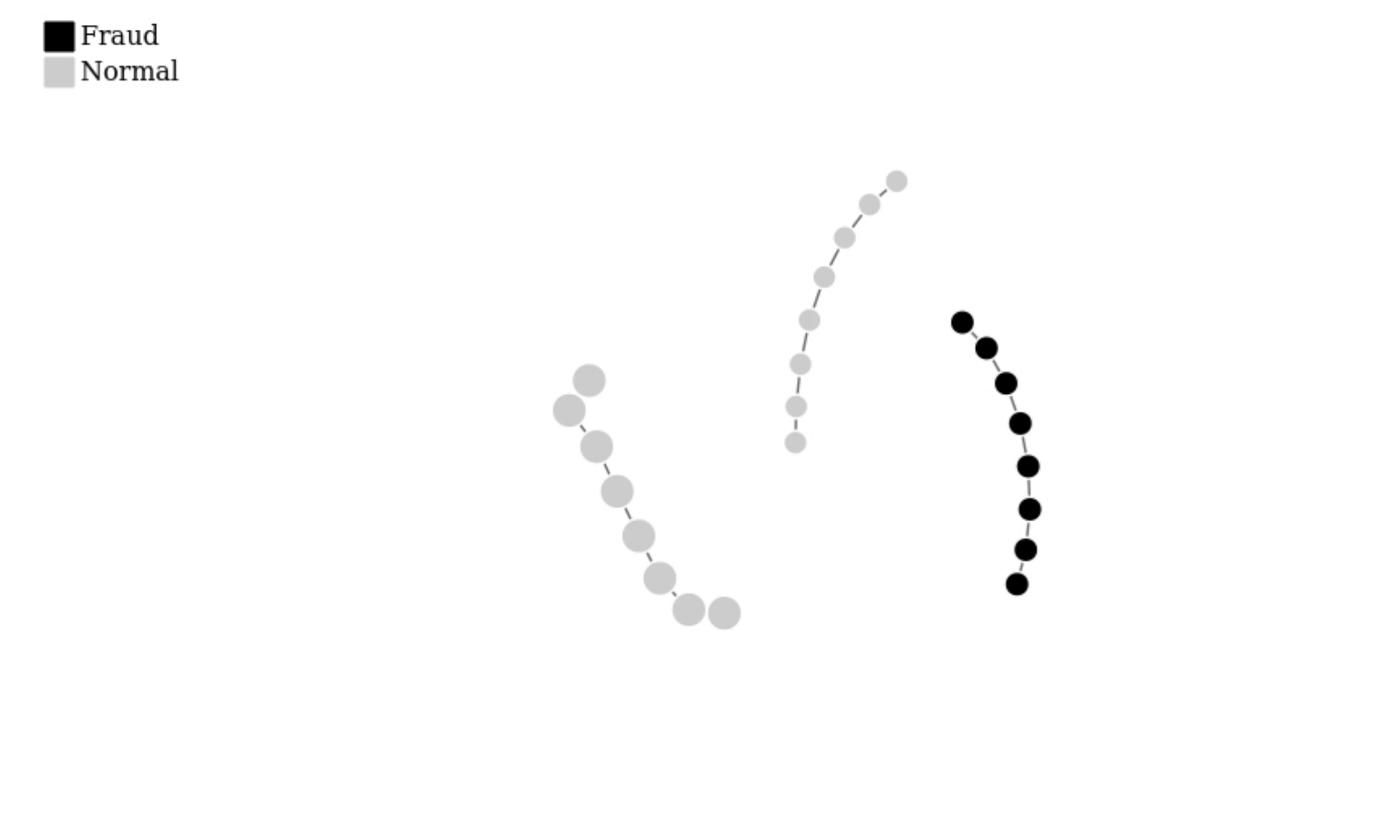}}
 \end{minipage}\hfill
 \begin{minipage}{.49\textwidth}
  \centering
  Reconstructed Samples \\ \vspace{0.2cm}
  \frame{\includegraphics[scale=0.235]{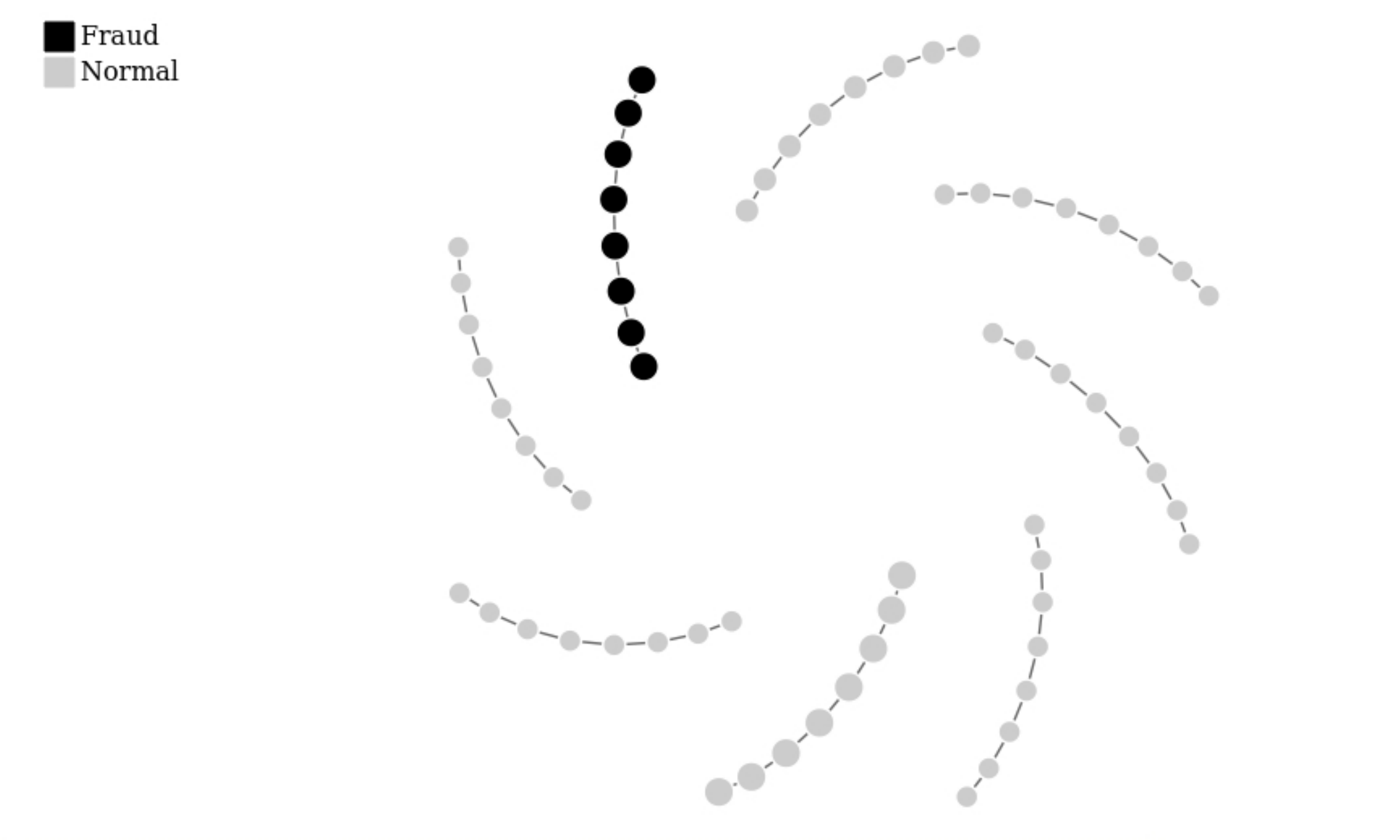}}
 \end{minipage}\hfill
 
 \begin{minipage}{.49\textwidth}
  \centering
  \vspace{0.2cm}
  \begin{turn}{90} 
   \hspace{0.5cm} PHom-VAE
  \end{turn}
  \frame{\includegraphics[scale=0.235]{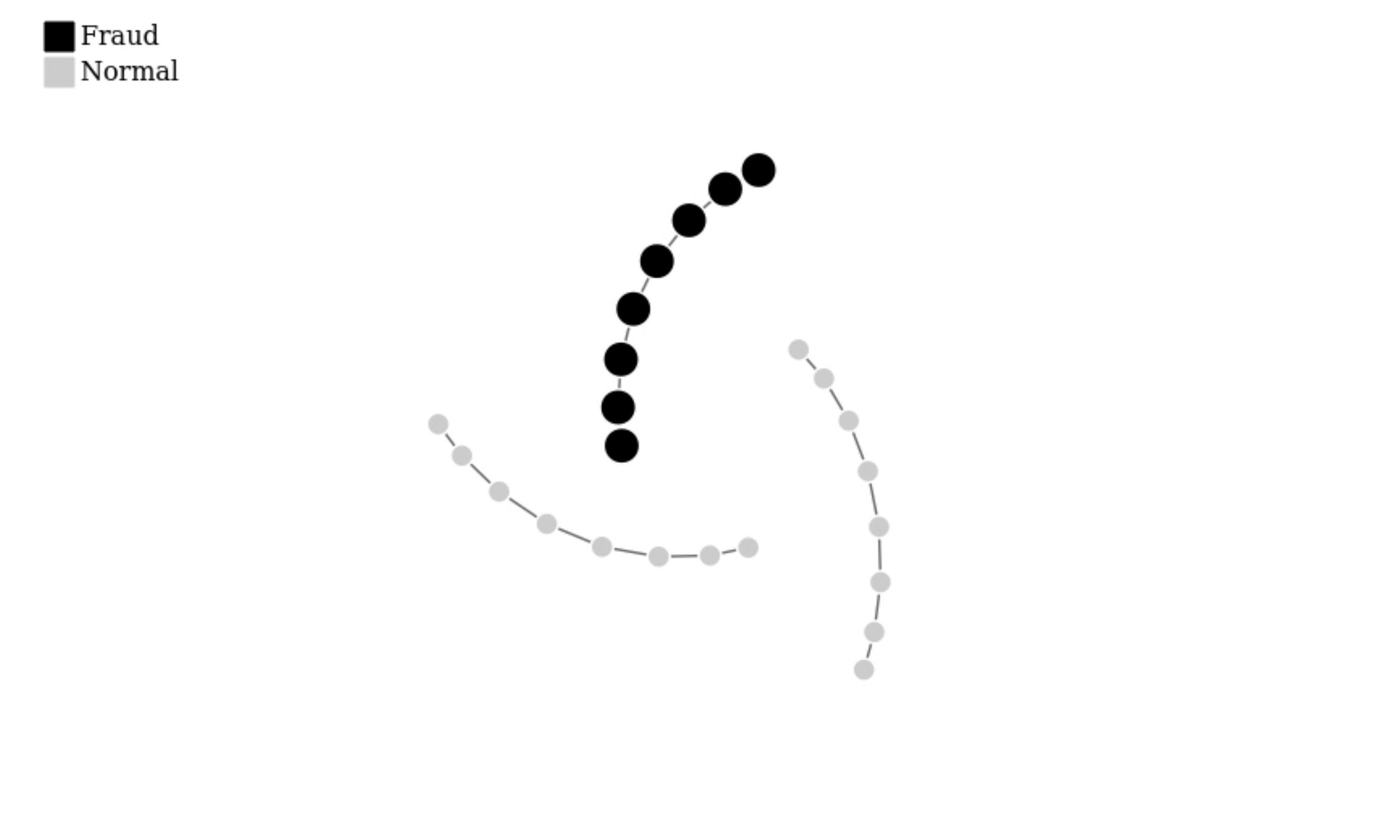}}
 \end{minipage}\hfill
 \begin{minipage}{.49\textwidth}
  \centering
  \vspace{0.2cm}
  \frame{\includegraphics[scale=0.235]{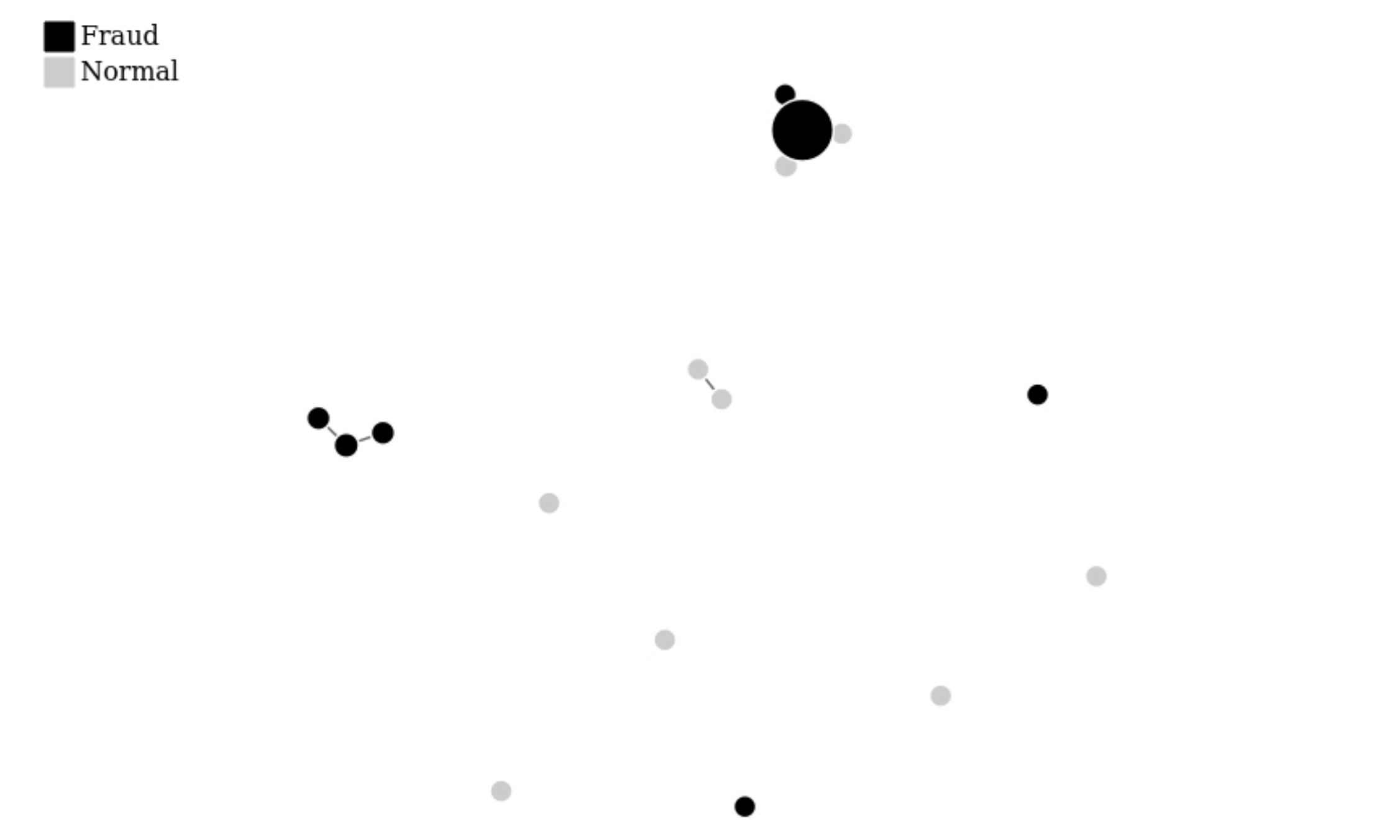}}
 \end{minipage}\hfill
 
\caption{Topological representation of the transaction samples, both fraudulent and normal, in black and gray respectively, of the reconstructed data distribution following $P_G(X|Z)$ for PHom-WAE and PHom-VAE. PHom-WAE does not alter the topological features of the manifold $\mathcal{X}$ through the encoding-decoding process.} \label{fig::reconstructedplot}
\end{figure*}

As pictured in figures \ref{fig::persistence_plot} and \ref{fig::latentplot}, %TopWAE is capable to represent more clusters in the topological representation of the latent manifold. TopWAE has longer inclusion maps and can differentiate normal and fraudulent transactions on a compressed latent manifold $\mathcal{Z}$ which is not the case of TopVAE. In figure \ref{fig::reconstructedplot}, the topological representation of the reconstructed samples $G(Z)$ of TopWAE is significantly closer to the original samples $X\in\mathcal{X}$ than TopVAE. TopWAE is capable of preserving the topological properties through the encoding-decoding process. It is illustrated by the fraudulent inclusion map separated from the normal inclusion maps in the reconstructed manifold $G(Z)$. The KL divergence $D_{KL}(P_X, P_G)$ used in TopVAE leads to a loss of the topological features. 
both the persistence diagram and the barcodes between the original and the reconstructed distributions, respectively $P_X$ and $P_G(X|Z)$, of the manifold $\mathcal{X}$ are more widely distributed for PHom-WAE than for PHom-VAE. Additionally, the persistence diagram and the barcodes of PHom-WAE are qualitatively closer to those associated with the original data manifold $\mathcal{X}$. It means the topological features are better preserved for PHom-WAE than for PHom-VAE. It highlights a better encoding-decoding process thanks to the use of an optimal transport cost function. Furthermore, in figure \ref{fig::reconstructedplot}, a topological representation of the original and the reconstructed distributions is highlighted. We observe the iterated inclusion chains are more similar for PHom-WAE than for PHom-VAE. For PHom-VAE, the inclusions of the reconstructed distribution are randomly scattered through the space without connected vertices. 

% \begin{figure*}[!p]

% \begin{figure*}[!p]

In order to quantitatively assess the quality of the encoding-decoding process, we use the bottleneck distance between the persistent diagram of $\mathcal{X}$ and the persistent diagram of $G(Z)$ of the reconstructed data points. We recall the strength of the bottleneck distance is to measure quantitatively the topological changes in the data, either the true or the reconstructed data, while being insensitive to the scale of the analysis. Traditional distance measures fail to acknowledge this as they do not rely on persistent homology and, therefore, can only reflect a measurement of the nearness relations of the data points without considering the overall shape of the data distribution. In table \ref{tab::res_ae}, we notice the smallest bottleneck distance, and therefore, the best result, is obtained with PHom-WAE. It means PHom-WAE is capable to better preserve the topological features of the original data distribution than PHom-VAE including the nearness measurements and the overall shape.  \\

\begin{table}[b!]
 \centering
 \caption{Bottleneck distance (smaller is better) for PHom-WAE and PHom-VAE between the samples $X$ of the original manifold $\mathcal{X}$ and the reconstructed manifold $G(Z|X)$ for $Z\in\mathcal{Z}$. Because of OT, PHom-WAE achieves better performance.}
 \label{tab::res_ae}
 \vspace{0.25cm}
 \begin{tabular}{cc|c}
  \toprule
  PHom-WAE \quad & \quad PHom-VAE \quad & \quad Difference (\%) \\
  \midrule
  \textbf{0.0788} & 0.0878 & 10.25  \\
  \bottomrule
 \end{tabular}
\end{table}

Last but not least, persistent homology and the bottleneck distance are used to highlight the scarcity of the distribution $P_Z$ of the latent manifold $\mathcal{Z}$. Using the bootstrapping technique of \cite{friedman2001elements}, we successively randomly select data samples $Z_i$ contained in the manifold $\mathcal{Z}$. The total number of selected samples $Z_i$ is at least 50\% of the total number of points contained in the manifold $\mathcal{Z}$ to ensure a reliable statistical representation. Assuming the data is not scattered, the bottleneck distance between the persistence diagrams of the samples $Z_i$ is small. On the opposite, if the data is chaotically scattered in $\mathcal{Z}$, then the topological features between the samples $Z_i$ are significantly different, and consequently, the bottleneck distance is large. In table \ref{tab::res_scarc}, the bottleneck distance is significantly lower for PHom-WAE than for PHom-VAE. Therefore, the level of scattered chaos for PHom-WAE is lower than for PHom-VAE. It also means the distribution $P_Z$ of the latent manifold $\mathcal{Z}$ is better topologically preserved for PHom-WAE than for PHom-VAE. The reconstructed distribution $P_G(X|Z)$ of $\mathcal{X}$ is, thus, less altered for PHom-WAE.  \\

\begin{table}[t]
 \centering
 \caption{Bottleneck distance (smaller is better) for PHom-WAE and PHom-VAE between the samples $Z_i$ of the latent manifold $\mathcal{Z}$ following $P_Z$ to detect scattered distribution. PHom-WAE better preserves the topological features during the encoding than PHom-VAE resulting in a manifold $\mathcal{Z}$ less chaotically scattered, highlighted by the smaller bottleneck distance.}
 \label{tab::res_scarc}
 \vspace{0.25cm}
 \begin{tabular}{cc|c}
  \toprule
  PHom-WAE \quad & \quad PHom-VAE \quad & \quad Difference (\%)  \\
  \midrule
  \textbf{0.0984} & 0.1372 &  28.28 \\
  \bottomrule
 \end{tabular}
\end{table}

% !!! NEW SECTION !!! %
% !!! =========== !!! %
\section{Conclusion} \label{sec::ccl}
Building upon WAE and VAE, we introduce PHom-WAE and PHom-VAE, a new characterization of the manifolds of the WAE and VAE, respectively, that uses topology and persistence homology to highlight the manifold properties and the scattered points of the hidden space. We discussed their relations with other AE modeling techniques. Furthermore, relying on persistence homology, the bottleneck distance has been introduced to estimate quantitatively the alteration of the topological features occurring during the encoding-decoding process, a specificity that current traditional distance measures fail to acknowledge. We conducted experiments showing the performance of PHom-WAE in comparison to PHom-VAE using a challenging imbalanced real-world open data set containing credit card transactions, particularly suitable for the banking industry. We showed the superior performance of PHom-WAE in comparison to PHom-VAE. Future work will include further exploration of the topological features such as the influence of the simplicial complex and the possibility to integrate a topological optimization function as a regularization term.

\bibliographystyle{./splncs04}
\bibliography{./zzz-mybibliography}
\end{document}